\pdfoutput=1

\documentclass[11pt]{article}

\usepackage[]{acl}

\usepackage{times}
\usepackage{latexsym}

\usepackage[T1]{fontenc}

\usepackage[utf8]{inputenc}

\usepackage{microtype}

\usepackage[switch]{lineno}
\usepackage{amssymb}
\usepackage{latexsym}
\usepackage{mathtools}
\usepackage{algorithm}
\usepackage[noend]{algpseudocode}
\usepackage{setspace}

\usepackage{graphicx}
\usepackage{amsmath}
\usepackage{amsthm}
\usepackage{amssymb}
\usepackage{booktabs}
\usepackage{graphicx} 
\usepackage{graphics}
\usepackage{amssymb}
\usepackage{amsmath}
\usepackage{pifont}
\usepackage{makecell}
\usepackage{resizegather}
\usepackage{etoolbox}
\usepackage{booktabs,makecell,tabularx} 
\usepackage{enumitem}
\usepackage{mathtools}
\usepackage{cleveref}
\usepackage{multirow}
\usepackage{booktabs,array}
\usepackage{amsmath, bm}
\usepackage{color, colortbl}
\usepackage{xcolor}
\usepackage{booktabs}
\usepackage{algorithm}
\usepackage{array}
\usepackage{subfigure}
\usepackage{multirow}
\usepackage{threeparttable}
\usepackage{tabularx}
\usepackage{booktabs}
\usepackage{colortbl}
\usepackage{xcolor}
\usepackage{stfloats}
\usepackage[noend]{algpseudocode}
\usepackage{mathtools}
\usepackage{makecell}
\usepackage{setspace}
\usepackage{resizegather}
\usepackage{etoolbox}
\usepackage{booktabs,makecell,tabularx} 
\usepackage{enumitem}
\usepackage{mathtools}
\usepackage{cleveref}
\usepackage{multirow}
\usepackage{booktabs,array}
\usepackage{amsmath, bm}

\newcolumntype{P}[1]{>{\centering\arraybackslash}p{#1}}

\let\oldnl\nl
\definecolor{Gray}{gray}{0.9}
\definecolor{LightCyan}{rgb}{0.88,1,1}
\newcommand{\nonl}{\renewcommand{\nl}{\let\nl\oldnl}}

\newcolumntype{H}{>{\setbox0=\hbox\bgroup}c<{\egroup}@{}}
\newcommand{\righttriangle}{\mathbin{\rotatebox[origin=c]{30}{$\text{\ding{115}}$}}}

\title{Text Adversarial Purification as Defense against Adversarial Attacks}

\author{Linyang Li\, Demin Song,  Xipeng Qiu \\
 School of Computer Science, Fudan University \\
 Shanghai Key Laboratory of Intelligent Information Processing, Fudan University \\
\{linyangli19, dmsong20, xpqiu\}@fudan.edu.cn
}

\begin{document}
\maketitle

\begin{abstract}

Adversarial purification is a successful defense mechanism against adversarial attacks without requiring knowledge of the form of the incoming attack.
Generally, adversarial purification aims to remove the adversarial perturbations therefore can make correct predictions based on the recovered clean samples.
Despite the success of adversarial purification in the computer vision field that incorporates generative models such as energy-based models and diffusion models,
using purification as a defense strategy against textual adversarial attacks is rarely explored.
In this work, we introduce a novel adversarial purification method that focuses on defending against textual adversarial attacks.
With the help of language models, we can inject noise by masking input texts and reconstructing the masked texts based on the masked language models.
In this way, we construct an adversarial purification process for textual models against the most widely used word-substitution adversarial attacks.
We test our proposed adversarial purification method on several strong adversarial attack methods including Textfooler and BERT-Attack and experimental results indicate that the purification algorithm can successfully defend against strong word-substitution attacks.

\end{abstract}

\section{Introduction}

Adversarial examples \cite{goodfellow2014explaining} can successfully mislead strong neural models in both computer vision tasks \cite{CarliniW16a} and language understanding tasks \cite{Alzantot,jin2019textfooler}.
An adversarial example is a maliciously crafted example attached with an imperceptible perturbation and can mislead neural networks. 
To defend attack examples of images, the most effective method is adversarial training \cite{goodfellow2014explaining,madry2019deep} which is a mini-max game used to incorporate perturbations into the training process.

Defending adversarial attacks is extremely important in improving model robustness. However, defending adversarial examples in natural languages is more challenging due to the discrete nature of texts. 
That is, gradients cannot be used directly in crafting perturbations.
The substitution-based adversarial examples are more complicated than gradient-based adversarial examples in images, making it difficult for neural networks to defend against these substitution-based attacks.

The first challenge of defending against adversarial attacks in NLP is that due to the discrete nature, these substitution-based adversarial examples can have substitutes in any token of the sentence and each substitute has a large candidate list. 
This would cause a combinatorial explosion problem, making it hard to apply adversarial training methods.
Strong attacking methods such as \citet{jin2019textfooler} show that using the crafted adversarial examples as data augmentation in adversarial training cannot effectively defend against these substitution-based attacks.
Further, defending strategies such as adversarial training rely on the assumption that the candidate lists of the substitutions are accessible.
However, the candidate lists of the substitutions should \textbf{not} be exposed to the target model; that is, the target model should be unfamiliar to the candidate list of the adversarial examples.
In real-world defense systems, the defender is not aware of the strategy the potential attacks might use, so the assumption that the candidate list is available would significantly constrain the potential applications of these defending methods.

Considering that it is challenging to defend against textual adversarial attacks when the form of the attacks cannot be acknowledged in advance, we introduce a novel adversarial purification method as a feasible defense mechanism against these attacks.
The adversarial purification method is to purify adversarially perturbed input samples before making predictions \cite{DBLP:journals/nn/SrinivasanRMMSN21,DBLP:conf/iclr/ShiHM21,DBLP:conf/icml/YoonHL21}.
The major works about adversarial purification focus on purifying continuous inputs such as images, therefore these works explore different generative models such as GANs \cite{Samangouei18defense}, energy-based models (EBMs) \cite{lecun2006tutorial} and recently developed diffusion models \cite{DBLP:conf/iclr/0011SKKEP21,DBLP:conf/icml/NieGHXVA22}.
However, in textual adversarial attacks, the inputs are discrete tokens which makes it more challenging to deploy previous adversarial purification methods.

Therefore, we introduce a purification mechanism with the help of masked language models.
We first consider the widely used masking process to inject noise into the input;
then we recover the clean texts from the noisy inputs with the help of the masked language models (e.g. a BERT \cite{bert}).
Further, considering that the iterative process in previous adversarial purification algorithms can be extremely costly (e.g. a VP-SDE process in diffusion models \cite{DBLP:conf/iclr/0011SKKEP21}), we instead simplify the iterative process to an ensemble-purifying process that conducting adversarial purification multiple times to obtain an ensembled result as a compromise to the time cost in traditional adversarial purification process.

Through extensive experiments, we prove that the proposed text adversarial purification algorithm can successfully serve as defense against strong attacks such as Textfooler and BERT-Attack.
Experiment results show that the accuracy under attack in baseline defense methods is lower than random guesses, while after text purification, the performance can reach only a few percent lower than the original accuracy when the candidate range of the attack is limited.
Further, extensive results indicate that the candidate range of the attacker score is essential for successful attacks, which is a key factor in maintaining the semantics of the adversaries.
Therefore we also recommend that future attacking methods can focus on achieving successful attacks with tighter constraints.

To summarize our contributions:

     (1) We raise the concern of defending substitution-based adversarial attacks without acknowledging the form of the attacks in NLP tasks.
     
     (2) To the best of our knowledge, we are the first to consider adversarial purification as a defense against textual adversarial attacks exemplified by strong word-substitution attacks and combine text adversarial purification with pre-trained models.
     
     (3) We perform extensive experiments to demonstrate that the adversarial purification method is capable of defending strong adversarial attacks, which brings a new perspective to defending textual adversarial attacks.

\section{Related Work}

\subsection{Adversarial Attacks in NLP}

In NLP tasks, current methods use substitution-based strategies \cite{Alzantot,jin2019textfooler,ren2019generating} to craft adversarial examples. 
Most works focus on the score-based black-box attack, that is, attacking methods know the logits of the output prediction.
These methods use different strategies \cite{Yoo2020SearchingFA,Morris2020ReevaluatingAE} to find words to replace, such as genetic algorithm \cite{Alzantot}, greedy-search \cite{jin2019textfooler,li2020bert} or gradient-based methods \cite{ebrahimi2017hotflip,cheng2019robust} and get substitutes using synonyms \cite{jin2019textfooler,mrkvsic2016counter,ren2019generating} or language models \cite{li2020bert,garg2020bae,shi2020robust}.

\subsection{Adversarial Defenses}

We divide the defense methods for word-substitution attacks by whether the defense method requires knowledge of the form of the attack.

When the candidate list is known, recent works introduce defense strategies that incorporate the candidates of the words to be replaced as an augmentation.
\citet{jin2019textfooler,li2020bert,si2020better} uses generated adversaries to augment the classifier for better defense performances;
\citet{jia2019certified,huang2019achieving} introduce a certified robust model to construct a certified space within the range of a candidate list therefore the substitutions in the candidate list cannot perturb the model.
\citet{zhou2020defense,dong2021towards} construct a convex hull based on the candidate list which can resist substitutions in the candidate list.

To defend unknown attacks, NLP models can incorporate gradient-based adversarial training strategies \cite{Miyato2016VirtualAT,madry2019deep} since recent works \cite{ebrahimi2017hotflip,cheng2019robust,zhu2019freelb,li2020textat} show that gradient-based adversarial training can also improve defense performances against word-substitution attacks.

\subsection{Adversarial Purification}

Adversarial purification is a defense strategy that uses generative models to purify adversarial inputs before making predictions, which is a promising direction in adversarial defense.
\citet{Samangouei18defense} uses a defensive GAN framework to build clean images to avoid adversarial attacks.
Energy-based models (EBMs) are used to purify attacked images via Langevin dynamics \cite{lecun2006tutorial}.
Score-based models \cite{Yoo2020SearchingFA} is also introduced as a purification strategy.
Recent works focus on exploring diffusion models as the purification model in purifying the attacked images \cite{DBLP:conf/icml/NieGHXVA22}.
Though widely explored, adversarial purification strategy is less explored in the NLP field.

\begin{figure*}[]
\centering
\includegraphics[width=0.95\linewidth]{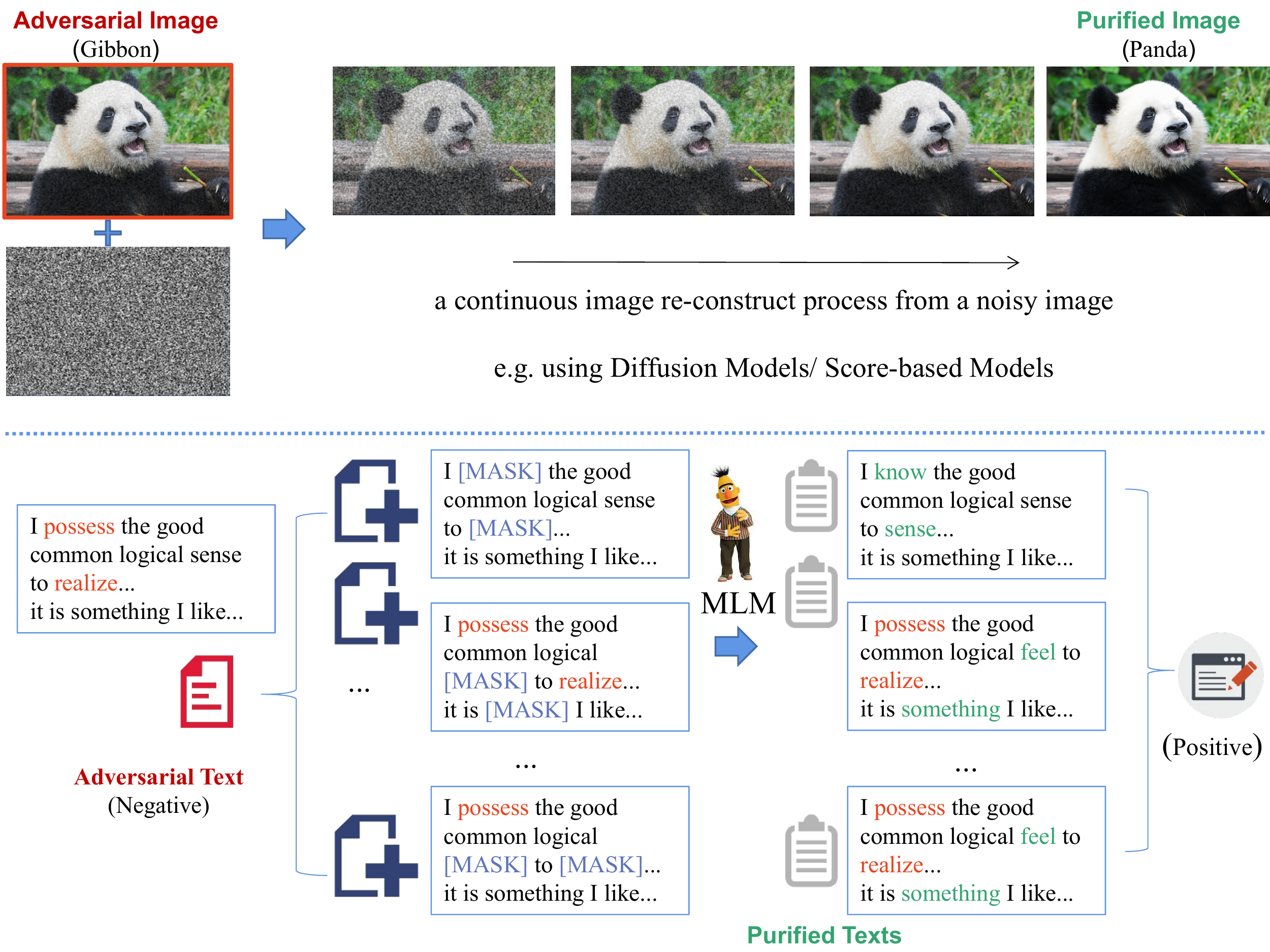}
\centering
\caption{Text Adversarial Purification Process: Compared with Image Purification, we use masked language models to recover noisy texts to purify adversarial texts as a defense against word-substitutions attacks.}
\label{fig:main}
\end{figure*}

\section{Text Adversarial Purification}

\subsection{Background of Adversarial Purification}

A classic adversarial purification process is to gradually purify the input through $T$ steps of purification runs.
As seen in Figure \ref{fig:main}, 
the purification process in the image domain is to first construct an input $x^{'}$ from the perturbed input $x$ by injecting random noise.
Then the purification algorithm will recover the clean image $\widehat{x}$ from the noisy image $x^{'}$ which usually takes multiple rounds.
The intuition of such a purification process is that the recovered inputs will not contain adversarial effects. 

Specifically, in the score-based adversarial purification \cite{Yoo2020SearchingFA},
the sample injected with random noise is
$x^{'} = x + \varepsilon $ where $\varepsilon \sim \mathcal{N}(0, \sigma^2I)$ and the goal is to purify $x^{'}$ with score network $s_{\theta}$.
In a continuous time step where $x_0 = x^{'}$, the goal is to recover $x_0$ through a score-based generative model $x_t = x_{t-1} + \alpha_{t-1} s_\theta(x_{t-1})$ where $\alpha$ is the step size related to $x_{t-1}$.
After $T$ times of generation, the recovered $\widehat{x} = x_T$ is used in the final prediction which contains less adversarial effect.

As for the diffusion-based purification methods \cite{DBLP:conf/icml/NieGHXVA22}, the process includes a forward diffusion process and a reverse recovery process.
The noise injection process is a forward stochastic differential equation (SDE), that is,
the noisy input $x^{'} = x(T)$ and initial perturbed input $x = x(0)$.
The diffusion process is $x(T) = \sqrt{\alpha(T)} x(0) + \sqrt{1-\alpha(T)} \varepsilon $ where $\alpha$ is a hyper-parameter and $\varepsilon \sim \mathcal{N}(0, \sigma^2I)$.
The final purified input $\widehat{x} = \widehat{x}(0)$ where $\widehat{x}(0)$ is the reverse-time SDE generated input from the diffused input $x(T)$.

\subsection{Text Adversarial Purification with BERT}

Instead of the iterative purification process used in purifying images, we introduce a novel purification method that purifies the input texts via masking and masks prediction with pre-trained masked language models exemplified by BERT \cite{bert}.

As seen in Figure \ref{fig:main}, instead of gradually adding noise and recovering the clean sample from the noisy samples,
we inject random noise into the input texts multiple times and recover the noisy data to a clean text based on the mask-prediction ability of the masked language model $F_m(\cdot)$.

Considering that the perturbed text is $X$, 
we can inject noise to construct multiple copies $X^{'}_i = [w_0, \cdots, \texttt{[MASK]}, w_n, \cdots, ] $.
We use two simple masking strategies: 
(1) Randomly mask the input texts;
(2) Randomly insert masks into the input texts.
Such a random masking process is similar to adding a random noise $\varepsilon \sim \mathcal{N}(0, \sigma^2I)$ to the inputs $x$.

After constructing multiple noisy inputs, we run the denoise process via masked language models:
$\widehat{X}_i = F_m(X^{'}_i) $.

With $N$ recovered texts, we are able to make predictions with the classifier $F_c(\cdot)$:
$S_i = \frac{1}{N} \sum_{i=0}^{N} \left( Softmax  (F^{}_{c}(\widehat{X}_i)) \right)$.

Unlike continuous perturbations to images, word-substitution adversarial samples only contain several perturbed words.
Therefore, we consider using a multiple-time mask-and-recover process as text adversarial purification, which makes full use of the pre-trained ability of the masked language models.
Compared with the generation process used in image adversarial purification,
masked language model-based purification method is easier to implement and utilize in pre-trained model-based applications as a defense against strong word-substitution adversarial attacks.

\subsection{Combining with Classifier}

Normal adversarial purification methods are plug-and-play processes inserted before the classification, 
however, the masked language model itself is a widely used classification model.
That is, the purification model $F_m(\cdot)$ and the classification model $F_c(\cdot)$ can share the same model.
Therefore, instead of using a normal masked language model such as BERT, we train the classifier and the mask-filling ability as multi-tasks.
The classification loss is $\mathcal{L}_{c} =   \mathcal{L}(F_{c}({X^{'}}),y, \theta) + \mathcal{L}(F_{c}({X}),y, \theta) $ and the masked language model loss is $\mathcal{L}_{mlm} =  \mathcal{L}(F_{m}({X^{'}}),X, \theta)$.
Here, the input $X$ is the clean text used in training the classifier and the $X^{'}$ is the random masked text.
The loss function $\mathcal{L}(\cdot)$ is the cross-entropy loss used in both the text classification head and masked language modeling head in the pre-trained models exemplified by BERT.

In this way, we are utilizing the pre-trained models to their full ability by using both the mask-filling function learned during the pre-training stage as well as the generalization ability to downstream tasks.

\begin{algorithm}[]
\setstretch{1.10}
\caption{Adversarial Training}\label{alg:rebuild-train}
\begin{algorithmic}[1]
\Require{Training Sample $X$, adversarial step $T_a$}
\State ${X^{'}} \gets$ Inject Noise $X$ 
\State $\boldsymbol{\delta}_{0}^{} \gets \frac{1}{\sqrt{D_{}}} \mathcal{N}(0, \sigma^2)$ \textcolor[rgb]{0.00,0.50,0.00}{// Init Perturb}
\For {$t = 0,1,... T_a$}

\State $\boldsymbol{g}_{\delta} \gets \bigtriangledown_{\delta}(\mathcal{L}_c + \mathcal{L}_{mlm}) $ \textcolor[rgb]{0.00,0.50,0.00}{// Get Perturbation}
\State $\boldsymbol{\delta}^{}_{t} \gets {\prod}_{{||\boldsymbol{\delta}_{}||}_F < \epsilon}(\boldsymbol{\delta}_{t}^{} + \alpha \cdot \boldsymbol{g}_{\delta}^{} / {||\boldsymbol{g}_{\delta}^{}||}_{F})$

\State $\mathcal{L}_{noise} \gets$  $\mathcal{L}_{}(F_{m}({X^{'}}+\boldsymbol{\delta}_t),X, \theta)$
\State ${X^{'}} \gets$ ${X^{'}} + \boldsymbol{\delta_t}$ \textcolor[rgb]{0.00,0.50,0.00}{// Update Input}

\State $\boldsymbol{g}_{t+1} = \boldsymbol{g}_{t} +  \bigtriangledown_{\theta}(\mathcal{L}_{c} + \mathcal{L}_{mlm} + \mathcal{L}_{noise})$ 
\EndFor
\State $\theta \gets \theta - \boldsymbol{g}_{T+1}$ \textcolor[rgb]{0.00,0.50,0.00}{// Update model parameter $\theta$}

\end{algorithmic}
\end{algorithm}

\subsection{Combining with Adversarial Training}

Different from the image field where adversaries are usually generated by gradients, 
word-substitution attacks do not have direct connections with gradient-based adversaries in the text domain.
Therefore, it is intuitive to incorporate gradient-based adversarial training in the purification process when the purification process is combined with the classifier training.

We introduce the adversarial training process therefore the purification function $F_m(\cdot)$ includes mask-prediction and recovering clean texts from inputs with gradient-based perturbations, which leads to stronger purification ability compared with a standard BERT.

Following standard adversarial training process with gradient-based adversaries introduced by \citet{zhu2019freelb,li2020textat}.
In the adversarial training process, a gradient-based perturbation $\boldsymbol{\delta}$ is added to the embedding output of the input text $X$ (for simplicity, we still use $X$ and $X^{'}$ to denote the embedding output in the Algorithm \ref{alg:rebuild-train}).
Then the perturbed inputs are added to the training set in the training process.
We combine gradient-based adversarial training with the text purification process.
As illustrated in Algorithm \ref{alg:rebuild-train}, for an adversarial training step, we add perturbations to the masked text $X^{'}$ and run $T_a$ times of updates.
We calculate gradients based on both classification losses $\mathcal{L}_c$ and masked language modeling losses $\mathcal{L}_{mlm}$;
further, as seen in line 6, we also calculate the loss that the masked language model will predict the texts from the perturbed text $X^{'} + \boldsymbol{\delta}$, which enhanced the text recover ability from noisy or adversarial texts.

\section{Experiments}

\subsection{Datasets}

We use two widely used text classification datasets: IMDB \footnote{https://datasets.imdbws.com/} \cite{maas2011learning} and AG's News \footnote{https://www.kaggle.com/amananandrai/ag-news-classification-dataset} \cite{zhang2015character} in our experiments.
The IMDB dataset is a bi-polar movie review classification task; the AG's News dataset is a four-class news genre classification task.
The average length is 220 words in the IMDB dataset, and 40 words in the AG's News dataset.
We use the test set following the Textfooler 1k test set in the main result and sample 100 samples for the rest of the experiments since the attacking process is seriously slowed down when the model is defensive.  

\subsection{Attack Methods}

Popular attack methods exemplified by genetic Algorithm \cite{Alzantot}, Textfooler \cite{jin2019textfooler} and BERT-Attack \cite{li2020bert} can successfully mislead strong models of both IMDB and AG's News task with a very small percentage of substitutions. 
Therefore, we use these strong adversarial attack methods as the attacker to test the effectiveness of our defense method.
The hyperparameters used in the attacking algorithm vary in different settings: we choose candidate list size $K$ to be 12, 48, and 50 which are used in the Textfooler and BERT-Attack methods.

We use the exact same metric used in Textfooler and BERT-Attack that calculates the after-attack accuracy, which is the targeted adversarial evaluation defined by \citet{si2020better}.
The after-attack accuracy measures the actual defense ability of the system under adversarial attacks.

\subsection{Victim Models and Defense Baselines}

The victim models are the fine-tuned pre-train models exemplified by BERT and RoBERTa, which we implement based on Huggingface Transformers \footnote{https://github.com/huggingface/transformers}
\cite{wolf-etal-2020-transformers}.
As discussed above, there are few works concerning adversarial defenses against attacks without knowing the candidates in NLP tasks.
Moreover, previous works do not focus on recent strong attack algorithms such as Textfooler \cite{jin2019textfooler}, BERT-involved attacks \cite{li2020bert,garg2020bae}
Therefore, we first list methods that can defend against adversarial attacks without accessing the candidate list as our baselines:

\textbf{Adv-Train (Adv-HotFlip)}: \citet{ebrahimi2017hotflip} introduces the adversarial training method used in defending against substitution-based adversarial attacks in NLP. 
It uses gradients to find actual adversaries in the embedding space.
    
\textbf{Virtual-Adv-Train (FreeLB)}: \citet{li2020textat,zhu2019freelb} use virtual adversaries to improve the performances in fine-tuning pre-trained models, which can also be used to deal with adversarial attacks without accessing the candidate list.
We follow the standard FreeLB training process to re-implement the defense results. 

Further, there are some works that require the candidate list, it is not a fair comparison with defense methods without accessing the candidates, so we list them separately:

\textbf{Adv-Augmentation}: We generate adversarial examples of the training dataset as a data augmentation method. 
We mix the generated adversarial examples and the original training dataset to train a model in a standard fine-tuning process. 

\textbf{ASCC}: \citet{dong2021towards} also uses a convex-hull concept based on the candidate vocabulary as a strong adversarial defense. 

\textbf{ADA}: \citet{si2020better} uses a mixup strategy based on the generated adversarial examples to achieve adversarial defense with variants AMDA-SMix that mixup the special tokens.

\textbf{FreeLB++}: \citet{li2021searching} introduces a variant of FreeLB method that expands the norm bound.

\textbf{RanMASK}: \citet{zeng2021certified} introduces a masking strategy that makes use of noises to improve robustness.

\begin{table*}[]
\centering

\scalebox{0.79}{
\begin{tabular}{l||c||cccc}
\toprule
\midrule
  \multirow{2}{*}{\textbf{Defense} $\downarrow$ \textbf{Attacks}$\rightarrow$}  & {\cellcolor{red!10} {Origin}} & \cellcolor{blue!10} Textfooler & \cellcolor{cyan!10} BERT-Attack & \cellcolor{blue!10} Textfooler & \cellcolor{cyan!10} BERT-Attack\\
  & \cellcolor{red!10} & \cellcolor{blue!10}(K=12)& \cellcolor{cyan!10}(K=12) & \cellcolor{blue!10}(K=50)& \cellcolor{cyan!10}(K=48)
  \\ \midrule

\midrule
\multicolumn{2}{l}{\textbf{IMDB} $\downarrow$}  \\
\midrule
 \rowcolor{gray!10}  BERT \cite{bert}  & 94.1 & 20.4 & 18.5 & 2.8 & 3.2   \\ 
 \rowcolor{gray!10}  RoBERTa \cite{liu2019roberta}& 97.3 & 26.3 & 24.5 & 25.2 & 23.0 \\ \midrule
 {\color{gray}\ding{108}} Adv-HotFlip (BERT) \cite{ebrahimi2017hotflip} & 95.1 & 36.1 & 34.2 & 8.0 & 6.2 \\

{\color{brown}\ding{110}} FreeLB (BERT) \cite{li2020textat} & 96.0 & 30.2 & 30.4 & 7.3 & 2.3 \\ 
{\color{brown}\ding{110}} FreeLB++ (BERT) \cite{li2021searching} & 93.2 & - & - & 45.3 & 39.9 \\
{\color[HTML]{7fc97f} \ding{115}} RanMASK (RoBERTa) \cite{zeng2021certified} & 93.0 & - & - & 23.7 & 26.8   \\

\midrule

{\color[HTML]{386cb0}$\righttriangle$} \underline{\textbf{Text Purification}(BERT)} & 93.0 & \textbf{81.5} & \textbf{76.7} & \textbf{51.0}  & \textbf{44.5} \\
{\color[HTML]{386cb0}$\righttriangle$} \underline{\textbf{Text Purification}(RoBERTa)} & 96.1 &  \textbf{84.2} & \textbf{82.0} & \textbf{54.3} & \textbf{52.2} \\

 \midrule
\multicolumn{2}{l}{\textbf{AG's News} $\downarrow$} \\
\midrule
 \rowcolor{gray!10}  BERT \cite{bert}  & 92.0 & 32.8 & 34.3 & 19.4 &  14.1 \\ 
 \rowcolor{gray!10}  RoBERTa \cite{liu2019roberta}& 97.3 & 26.3 & 24.5 & 25.2 & 23.0 \\ 
 \midrule
 {\color{gray}\ding{108}} Adv-HotFlip (BERT) & 91.2 & 35.3 & 34.1 & 18.2 & 8.5\\
{\color{brown}\ding{110}} FreeLB (BERT) & 90.5 & 40.1 & 34.2 & 20.1 & 8.5\\
\midrule
{\color[HTML]{386cb0}$\righttriangle$} \underline{\textbf{Text Purification}(BERT)} & 90.6 & \textbf{61.5} & \textbf{49.7} & \textbf{34.9} & \textbf{22.5} \\
{\color[HTML]{386cb0}$\righttriangle$} \underline{\textbf{Text Purification}(RoBERTa)} & 90.8 & \textbf{59.1} & \textbf{41.2} & \textbf{34.2} & \textbf{19.5} \\
\bottomrule
\end{tabular}
}
\caption{After-Attack Accuracy compared with defense methods that can defend attacks without acknowledging the form of the attacks. That is, the substitution candidates of the attack methods are unknown to defense systems.
}
\label{tab:main-results}
\end{table*}

\subsection{Implementations}

We use BERT-BASE and RoBERTa-BASE models based on the Huggingface Transformers \footnote{https://github.com/huggingface/transformers}.
We modify the adversarial training with virtual adversaries based on the implementation of FreeLB, TAVAT, and FreeLB++.
The training hyper-parameters we use are different from FreeLB and TAVAT since we aim to find large perturbations to simulate adversaries.
We set adversarial learning rate $\alpha=$  1e-1 to and normalization boundary $\epsilon=$ 2e-1 in all tasks. 
We set the multiple purification size $N=$ to 16 for all tasks and we will discuss the selection of $N$ in the later section.

\begin{table}[]
\setlength{\tabcolsep}{2pt}
\centering
\small
\begin{tabular}{l||c||cccc}
\toprule

Methods & {\cellcolor{red!10}} Origin & \cellcolor{blue!10}  Textfooler & \cellcolor{cyan!10}  GA  \\
\midrule
\multicolumn{1}{l}{\bfseries IMDB $\downarrow$}\\
\midrule

 \rowcolor{gray!10} BERT  & 94.0 & 2.0 & 45.0 \\
{\color{gray}\ding{110}} Data-Augmentation & 93.0 & 18.0 & 53.0 \\
 {\color{brown}\ding{108}}ADA \cite{si2020better} & 96.7 & 3.0 & - \\
 {\color{brown}\ding{108}}AMDA\cite{si2020better} &  96.9 & 17.4 & -\\
{\color{yellow} \ding{115}} ASCC \cite{dong2021towards} & 77.0 & - & 71.0 \\
\midrule
{\color[HTML]{386cb0}$\righttriangle$} \underline{\textbf{Text Purification}(BERT)} & 93.0 & \textbf{51.0} & \textbf{79.0} \\

\bottomrule

\end{tabular}
\caption{After-Attack Accuracy compared with access-candidates methods based on the BERT model.
Here we implement Textfooler with K=50 for consistency with previous works.
GA is the Genetic Attack method.
We use the AMDA-SMix setup for the AMDA method.
}
\label{tab:known-results}
\end{table}

For our text adversarial purification method, we use the model that is trained with gradient-based adversarial training as the purification model $F_m(\cdot)$ and the classifier $F_c(\cdot)$ for the main experiments and conduct thorough ablations to explore the effect of combining purification with classifier and adversarially trained classifier.

As for implementing adversarial attack methods, we use the TextAttack toolkit while referring the official codes of the corresponding attack methods \footnote{https://github.com/QData/TextAttack} \cite{morris2020textattack}.
The similarity thresholds of the word-substitution range are the main factors of the attacking algorithm.
We tune the USE \cite{cer2018universal} constraint 0.5 for the AG task and 0.7 for the IMDB task and 0.5 for the cosine-similarity threshold of the synonyms embedding \cite{mrkvsic2016counter} which can reproduce the results of the attacking methods reported.

\begin{table*}[]
\centering

\scalebox{0.79}{
\begin{tabular}{l||c||cccc}
\toprule
\midrule
  \multirow{2}{*}{\textbf{Defense} $\downarrow$ \textbf{Attacks}$\rightarrow$}  & {\cellcolor{red!10} {Origin}} & \cellcolor{blue!10} Textfooler & \cellcolor{cyan!10} BERT-Attack \\
  & \cellcolor{red!10} & \cellcolor{blue!10}(K=12)& \cellcolor{cyan!10}(K=12) 
  \\ \midrule

\midrule
\multicolumn{2}{l}{{\color[HTML]{386cb0}$\righttriangle$} \textbf{Text Purification Only} $\downarrow$}  \\
\midrule
{\color[HTML]{eb9634}\ding{52}} Purification & 94.0 & 72.0 & 60.0  \\ 
{\color[HTML]{eb9634}\ding{52}} Purification {\color[HTML]{00994c}\ding{54}} Multi. Recovery & 87.0 & 20.0 & 13.0 \\ 
{\color[HTML]{eb9634}\ding{52}} Purification {\color[HTML]{00994c}\ding{54}} Mask Insertion {\color[HTML]{00994c}\ding{54}} Multi. Recovery & 92.0 & 11.0 & 3.0\\
\midrule
\multicolumn{2}{l}{{\color[HTML]{386cb0}$\righttriangle$} \textbf{Combining Classifier} $\downarrow$}  \\
\midrule

{\color[HTML]{eb9634}\ding{52}}Purification {\color[HTML]{eb9634}\ding{52}} Comb. Classifier & 95.0 & 76.0 & 67.0 \\ 
{\color[HTML]{eb9634}\ding{52}}Purification {\color[HTML]{eb9634}\ding{52}} Comb. Classifier {\color[HTML]{00994c}\ding{54}} Multi. Recovery & 95.0 & 45.0 & 34.0 \\ 
{\color[HTML]{eb9634}\ding{52}}Purification {\color[HTML]{eb9634}\ding{52}} Comb. Classifier {\color[HTML]{00994c}\ding{54}} Multi. Recovery {\color[HTML]{00994c}\ding{54}} Mask Insertion & 95.0 & 29.0 & 17.0  \\

\midrule
\multicolumn{2}{l}{{\color[HTML]{386cb0}$\righttriangle$} \textbf{Combining Adversarially Trained Classifier} $\downarrow$}  \\
\midrule
{\color[HTML]{eb9634}\ding{52}} Purification {\color[HTML]{eb9634}\ding{52}} AT Classifier & 93.0 & \textbf{86.0} & \textbf{77.0}\\ 
{\color[HTML]{eb9634}\ding{52}} Purification {\color[HTML]{eb9634}\ding{52}} AT Classifier {\color[HTML]{00994c}\ding{54}} Multi. Recovery & 93.0 & 63.0 & 52.0 \\ 
{\color[HTML]{eb9634}\ding{52}} Purification {\color[HTML]{eb9634}\ding{52}} AT Classifier {\color[HTML]{00994c}\ding{54}} Multi. Recovery {\color[HTML]{00994c}\ding{54}} Mask  Insertion & 93.0 & 42.0 &29.0 \\
\midrule

 \rowcolor{gray!10} BERT & 94.0 & 10.0 & 5.0 \\

\bottomrule
\end{tabular}
}
\caption{Ablations results tested on attacking the IMDB task based on BERT models. Comb. Classifier is the combined fine-tuned $F_c(\cdot)$ and $F_m(\cdot)$ and AT Classifier is the adversarially trained $F_c(\cdot)$.
Mask Insertion is to use both mask-replacing and mask-insertion in injecting noise.
}
\label{tab:ablation-results}
\end{table*}

\subsection{Results}

As seen in Table \ref{tab:main-results}, the proposed \textbf{Text Adversarial Purification} algorithm can successfully defend strong attack methods.
The accuracy of our defending method under attack is significantly higher than non-defense models (50\% vs 20\% in the IMDB dataset).
Compared with previous defense methods, our proposed method can achieve higher defense accuracy in both the IMDB task and AG's News task.
The Adv-HotFlip and the FreeLB methods are effective, which indicates that gradient-based adversaries are not very similar to actual substitutions.
We can see that Adv-HotFlip and FreeLB methods achieve similar results (around 30\% when $K=12$) which indicates that gradient-based adversarial training methods have similar defense abilities no matter whether the adversaries are virtual or real since they are both unaware of the attacker's candidate list.
Also, the original accuracy (on the clean data) of our method is only a little lower than the baseline methods, which indicates that the purified texts still contain enough information for classification. 
The RoBERTa model also shows robustness using both original fine-tuned model and our defensive framework, which indicates our purification algorithm can be used in various pre-trained language models.
Compared with methods that specifically focus on adversarial defense, our proposed method can still surpass the state-of-the-art defense system FreeLB++ \cite{li2021searching} and RanMASK \cite{zeng2021certified}.

Further, the candidate size is extremely important in defending against adversarial attacks, when the candidate size is smaller, exemplified by $K=12$, our method can achieve very promising results.
As pointed out by \citet{Morris2020ReevaluatingAE}, the candidate size should not be too large that the quality of the adversarial examples is largely damaged.

As seen in Table \ref{tab:known-results}, we compare our method with previous access-candidates defense methods.
When defending against the widely used Textfooler attack and genetic attack \cite{Alzantot}, our method can achieve similar accuracy even compared with known-candidates defense methods.
As seen, the data augmentation method cannot significantly improve model robustness since the candidates can be very diversified. 
Therefore, using generated adversarial samples as an augmentation strategy does not guarantee robustness against greedy-searched methods like Textfooler and BERT-Attack.

\subsection{Analysis}

\subsubsection{Ablations}

As we design an adversarial purification algorithm with masked language models and propose a multiple-recovering strategy, we aim to explore which process helps more in the purification defense system.
Plus, we combine classifiers within the purification model so it is also important to explore whether such a combination is helpful.

For each type of purification method, we test whether the specific purification process we propose is effective.
That is, we test whether making multiple recoveries in the purification process is helpful;
also, we test whether using both masking tokens and inserting additional masks is helpful.

As seen in Table \ref{tab:ablation-results}, we can summarize that:

(1) Multi-time recovering is necessary: in the image domain, multiple reconstructions with a continuous time purification process are necessary. Similarly, the multi-recovery process is important in obtaining high-quality purification results.
We can observe that one-time recovery cannot achieve promising defense performances.

(2) Combining classifiers is effective: we can observe that when we use trained classifiers and masked language models, the defense performances are better than using fine-tuned classifier and vanilla BERT as a masked language model, indicating that such a combined training process is helpful in obtaining more strong defense systems.
Also, with gradient-based adversarial training, the purification process can obtain a further boost, indicating that our proposed text purification algorithm can be used together with previous defense methods as an advanced defense system.

\begin{table*}[]
    \setlength{\tabcolsep}{6pt}
    \scriptsize
    \centering
    \begin{tabular}{lll}
    \toprule
    \multirow{2}{*}{} & \bfseries{Texts} & Confidence \\
    & & \textcolor{blue}{(Positive)} \\
    \midrule
    Clean-Sample & \makecell[l]{
    I have the good common logical sense to know that oil can not last forever and I am acutely\\
    aware of how much of my life in the suburbs revolves around petrochemical products. I've \\
    been an avid consumer of new technology and I keep running out of space on powerboards - so... \\
    } & \textcolor{blue}{93.2\%}\\
    \midrule
    
    Adv. of BERT & \makecell[l]{
    I \textcolor{red}{possess} the good common logical sense to \textcolor{red}{realize} that oil can not last forever and I am acutely\\
    aware of how much of my life in the suburbs \textcolor{red}{spins} around petrochemical products. I've \\
    been an avid consumer of new technology and I keep running out of space on powerboards - \textcolor{red}{well}... \\
    } & \textcolor{red}{38.3\%}\\
    \midrule
    
    Adv. of Text Pure & \makecell[l]{
    I \textcolor{red}{know} the \textcolor{red}{wonderful} \textcolor{red}{general} sense to \textcolor{red}{knows} that \textcolor{red}{oils} can not last \textcolor{red}{endless} and I am acutely\\
    \textcolor{red}{know} of how \textcolor{red}{majority} of my \textcolor{red}{lived} in the \textcolor{red}{city} \textcolor{red}{spins} around petrochemical products . I've \\
    been an \textcolor{red}{amateur} \textcolor{red}{consumers} of \textcolor{red}{newly} \textcolor{red}{technologies} and I \textcolor{red}{kept} \textcolor{red}{working} out of \textcolor{red}{spaces} on powerboards \textcolor{red}{!} \textcolor{red}{well}... \\
    } & \textcolor{blue}{80.1\%} \\

    \hline
    \multirow{4}{*}{Purified Texts} & \makecell[l]{
    \textcolor{teal}{Well} I know the wonderful general sense notion to knows that oils \textcolor{teal}{production} can not last \textcolor{teal}{for} endless \textcolor{teal}{years} and I am  \\
    acutely know of how the majority of my live in the city \textcolor{red}{spins} around \textcolor{teal}{the} petrochemical production ... I've\\
    been an amateur consumers of new technologies and I kept working out of spaces on \textcolor{teal}{power skateboards}! \textcolor{red}{well} ...
    } & \textcolor{blue}{80.4\%}\\

    \cline{2-2}
    & \makecell[l]{
    I know the wonderful \textcolor{teal}{common} sense notion to knows that oils can not last \textcolor{teal}{forever} and I \textcolor{teal}{also} acutely know\\
     of how majority of my lived in the \textcolor{teal}{world} \textcolor{teal}{and} around petrochemical production ... I've\\
    been an amateur consumers of newly technologies and I kept working out of \textcolor{teal}{them} on \textcolor{teal}{skateboards} ! \textcolor{red}{well} ...    
    }& \textcolor{blue}{81.4\%} \\

    \cline{2-2}
    & \makecell[l]{
    I know the \textcolor{teal}{wonderfully} general sense notion to knows that oils can not last endless and I am acutely know\\
     of how majority \textcolor{teal}{part} of my lived in the \textcolor{teal}{big} city \textcolor{red}{spins} around \textcolor{teal}{petrocochemical} production ... I \textcolor{teal}{should have}\\
    been an amateur consumers \textcolor{teal}{fan} of newly technologies and I kept \textcolor{teal}{on} working out of spaces \textcolor{teal}{and} on powerboards ! \textcolor{red}{well} ...
    } & \textcolor{blue}{76.2\%} \\

    \cline{2-2}
    & \makecell[l]{
    I \textcolor{teal}{am} the \textcolor{teal}{the} general sense notion \textcolor{teal}{and} knows that oils can not last endless and I am acutely \\
    know of \textcolor{teal}{the} \textcolor{teal}{part} of my lived \textcolor{teal}{as} the city \textcolor{red}{spins} around petrochemical production ... I've\\
    been an amateur consumers of newly technologies and I kept working out of \textcolor{teal}{bed} on powerboards ! \textcolor{red}{well} ...
    } & \textcolor{blue}{78.5\%} \\

    \bottomrule
    \end{tabular}
    \caption{A random selected sample that BERT model failed to defend against the Textfooler Attack in the IMDB dataset and Text Pure (Text Adversarial Purification) succeed.
    Adv. of BERT is the adversarial sample generated by Textfooler to attack the classifier.
    Adv. of Text Pure is the sample generated by Textfooler to attack the classifier but failed.
    The purified texts are also listed.
    }
    \label{tab:erroanalysis}
\end{table*}

\subsubsection{Example of Purification Results}

As seen in Table \ref{tab:erroanalysis}, we construct multiple recoveries and use the averaged score as the final classification result.
Such a purification process is effective compared with vanilla fine-tuned BERT.

We can observe that the adversarial sample that successfully attacked the vanilla BERT model only achieves this by replacing only a few tokens.
While with the purification process, the attack algorithm is struggling in finding effective substitutions to achieve a successful attack.
Even replacing a large number of tokens that seriously hurt the semantics of the input texts, with the purification process involved, the classifier can still resist the adversarial effect.
Further, by observing the purified texts, we can find that the purified texts can make predictions correctly though some substitutes still exist in the purified texts, indicating that making predictions based on purified texts using the combined trained classifier can obtain a promising defense performance.
That is, our proposed method, though is not a plug-and-play system, can be used as a general system as a defense against substitution-based attacks.

\section{Conclusion and Future Work}

In this paper, we introduce a textual adversarial purification algorithm as a defense against substitution-based adversarial attacks.
We utilize the mask-infill ability of pre-trained models to recover noisy texts and use these purified texts to make predictions.
Experiments show that the purification method is effective in defending strong adversarial attacks without acknowledging the substitution range of the attacks.
We are the first to consider the adversarial purification method with a multiple-recovering strategy in the text domain while previous successes of adversarial purification strategies usually focus on the image field.
Therefore, we hope that the adversarial purification method can be further explored in NLP applications as a powerful defense strategy.

\clearpage

\section*{Limitations}

In this paper, we discuss an important topic in the NLP field, the defense against adversarial attacks in NLP applications.
We provide a strong defense strategy against the most widely used word substitution attacks in the NLP field, which is limited in several directions. 

\begin{itemize}
    \item We are testing defense strategies using downstream task models such as BERT and RoBERTa, and the purification tool is a model with a mask-filling ability such as BERT.
    Such a process can be further improved with strong models such as large language models.
    \item We study the concept of adversarial purification in the adversarial attack scenarios with word-substitution attacks on small fine-tuned models.
    The concept of adversarial purification can be further expanded to various NLP applications.
    For instance, the purification of natural language can be used in malicious text purification which is more suitable in applications with large language models.
    
\end{itemize}




\bibliography{anthology}

\begin{thebibliography}{38}
\expandafter\ifx\csname natexlab\endcsname\relax\def\natexlab#1{#1}\fi

\bibitem[{Alzantot et~al.(2018)Alzantot, Sharma, Elgohary, Ho, Srivastava, and
  Chang}]{Alzantot}
Moustafa Alzantot, Yash Sharma, Ahmed Elgohary, Bo{-}Jhang Ho, Mani~B.
  Srivastava, and Kai{-}Wei Chang. 2018.
\newblock \href {http://arxiv.org/abs/1804.07998} {Generating natural language
  adversarial examples}.
\newblock \emph{CoRR}, abs/1804.07998.

\bibitem[{Carlini and Wagner(2016)}]{CarliniW16a}
Nicholas Carlini and David~A. Wagner. 2016.
\newblock \href {http://arxiv.org/abs/1608.04644} {Towards evaluating the
  robustness of neural networks}.
\newblock \emph{CoRR}, abs/1608.04644.

\bibitem[{Cer et~al.(2018)Cer, Yang, Kong, Hua, Limtiaco, John, Constant,
  Guajardo-Cespedes, Yuan, Tar et~al.}]{cer2018universal}
Daniel Cer, Yinfei Yang, Sheng-yi Kong, Nan Hua, Nicole Limtiaco, Rhomni~St
  John, Noah Constant, Mario Guajardo-Cespedes, Steve Yuan, Chris Tar, et~al.
  2018.
\newblock Universal sentence encoder.
\newblock \emph{arXiv preprint arXiv:1803.11175}.

\bibitem[{Cheng et~al.(2019)Cheng, Jiang, and Macherey}]{cheng2019robust}
Yong Cheng, Lu~Jiang, and Wolfgang Macherey. 2019.
\newblock Robust neural machine translation with doubly adversarial inputs.
\newblock \emph{arXiv preprint arXiv:1906.02443}.

\bibitem[{Devlin et~al.(2018)Devlin, Chang, Lee, and Toutanova}]{bert}
Jacob Devlin, Ming{-}Wei Chang, Kenton Lee, and Kristina Toutanova. 2018.
\newblock \href {http://arxiv.org/abs/1810.04805} {{BERT:} pre-training of deep
  bidirectional transformers for language understanding}.
\newblock \emph{CoRR}, abs/1810.04805.

\bibitem[{Dong et~al.(2021)Dong, Liu, Ji, and Luu}]{dong2021towards}
Xinshuai Dong, Hong Liu, Rongrong Ji, and Anh~Tuan Luu. 2021.
\newblock \href {https://openreview.net/forum?id=ks5nebunVn_} {Towards
  robustness against natural language word substitutions}.
\newblock In \emph{International Conference on Learning Representations}.

\bibitem[{Ebrahimi et~al.(2017)Ebrahimi, Rao, Lowd, and
  Dou}]{ebrahimi2017hotflip}
Javid Ebrahimi, Anyi Rao, Daniel Lowd, and Dejing Dou. 2017.
\newblock Hotflip: White-box adversarial examples for text classification.
\newblock \emph{arXiv preprint arXiv:1712.06751}.

\bibitem[{Garg and Ramakrishnan(2020)}]{garg2020bae}
Siddhant Garg and Goutham Ramakrishnan. 2020.
\newblock Bae: Bert-based adversarial examples for text classification.
\newblock \emph{arXiv preprint arXiv:2004.01970}.

\bibitem[{Goodfellow et~al.(2014)Goodfellow, Shlens, and
  Szegedy}]{goodfellow2014explaining}
Ian~J Goodfellow, Jonathon Shlens, and Christian Szegedy. 2014.
\newblock Explaining and harnessing adversarial examples.
\newblock \emph{arXiv preprint arXiv:1412.6572}.

\bibitem[{Huang et~al.(2019)Huang, Stanforth, Welbl, Dyer, Yogatama, Gowal,
  Dvijotham, and Kohli}]{huang2019achieving}
Po-Sen Huang, Robert Stanforth, Johannes Welbl, Chris Dyer, Dani Yogatama, Sven
  Gowal, Krishnamurthy Dvijotham, and Pushmeet Kohli. 2019.
\newblock Achieving verified robustness to symbol substitutions via interval
  bound propagation.
\newblock \emph{arXiv preprint arXiv:1909.01492}.

\bibitem[{Jia et~al.(2019)Jia, Raghunathan, G{\"{o}}ksel, and
  Liang}]{jia2019certified}
Robin Jia, Aditi Raghunathan, Kerem G{\"{o}}ksel, and Percy Liang. 2019.
\newblock \href {http://arxiv.org/abs/1909.00986} {Certified robustness to
  adversarial word substitutions}.
\newblock \emph{CoRR}, abs/1909.00986.

\bibitem[{Jin et~al.(2019)Jin, Jin, Zhou, and Szolovits}]{jin2019textfooler}
Di~Jin, Zhijing Jin, Joey~Tianyi Zhou, and Peter Szolovits. 2019.
\newblock \href {http://arxiv.org/abs/1907.11932} {Is {BERT} really robust?
  natural language attack on text classification and entailment}.
\newblock \emph{CoRR}, abs/1907.11932.

\bibitem[{LeCun et~al.(2006)LeCun, Chopra, Hadsell, Ranzato, and
  Huang}]{lecun2006tutorial}
Yann LeCun, Sumit Chopra, Raia Hadsell, M~Ranzato, and F~Huang. 2006.
\newblock A tutorial on energy-based learning.
\newblock \emph{Predicting structured data}, 1(0).

\bibitem[{Li et~al.(2020)Li, Ma, Guo, Xue, and Qiu}]{li2020bert}
Linyang Li, Ruotian Ma, Qipeng Guo, Xiangyang Xue, and Xipeng Qiu. 2020.
\newblock Bert-attack: Adversarial attack against bert using bert.
\newblock \emph{arXiv preprint arXiv:2004.09984}.

\bibitem[{Li and Qiu(2020)}]{li2020textat}
Linyang Li and Xipeng Qiu. 2020.
\newblock Textat: Adversarial training for natural language understanding with
  token-level perturbation.
\newblock \emph{arXiv preprint arXiv:2004.14543}.

\bibitem[{Li et~al.(2021)Li, Xu, Zeng, Li, Zheng, Zhang, Chang, and
  Hsieh}]{li2021searching}
Zongyi Li, Jianhan Xu, Jiehang Zeng, Linyang Li, Xiaoqing Zheng, Qi~Zhang,
  Kai-Wei Chang, and Cho-Jui Hsieh. 2021.
\newblock Searching for an effective defender: Benchmarking defense against
  adversarial word substitution.
\newblock \emph{arXiv preprint arXiv:2108.12777}.

\bibitem[{Liu et~al.(2019)Liu, Ott, Goyal, Du, Joshi, Chen, Levy, Lewis,
  Zettlemoyer, and Stoyanov}]{liu2019roberta}
Yinhan Liu, Myle Ott, Naman Goyal, Jingfei Du, Mandar Joshi, Danqi Chen, Omer
  Levy, Mike Lewis, Luke Zettlemoyer, and Veselin Stoyanov. 2019.
\newblock Roberta: A robustly optimized bert pretraining approach.
\newblock \emph{arXiv preprint arXiv:1907.11692}.

\bibitem[{Maas et~al.(2011)Maas, Daly, Pham, Huang, Ng, and
  Potts}]{maas2011learning}
Andrew Maas, Raymond~E Daly, Peter~T Pham, Dan Huang, Andrew~Y Ng, and
  Christopher Potts. 2011.
\newblock Learning word vectors for sentiment analysis.
\newblock In \emph{Proceedings of the 49th annual meeting of the association
  for computational linguistics: Human language technologies}, pages 142--150.

\bibitem[{Madry et~al.(2019)Madry, Makelov, Schmidt, Tsipras, and
  Vladu}]{madry2019deep}
Aleksander Madry, Aleksandar Makelov, Ludwig Schmidt, Dimitris Tsipras, and
  Adrian Vladu. 2019.
\newblock \href {http://arxiv.org/abs/1706.06083} {Towards deep learning models
  resistant to adversarial attacks}.

\bibitem[{Miyato et~al.(2016)Miyato, Dai, and Goodfellow}]{Miyato2016VirtualAT}
Takeru Miyato, Andrew~M. Dai, and Ian~J. Goodfellow. 2016.
\newblock Virtual adversarial training for semi-supervised text classification.
\newblock \emph{ArXiv}, abs/1605.07725.

\bibitem[{Morris et~al.(2020{\natexlab{a}})Morris, Lifland, Yoo, Grigsby, Jin,
  and Qi}]{morris2020textattack}
John Morris, Eli Lifland, Jin~Yong Yoo, Jake Grigsby, Di~Jin, and Yanjun Qi.
  2020{\natexlab{a}}.
\newblock Textattack: A framework for adversarial attacks, data augmentation,
  and adversarial training in nlp.
\newblock In \emph{Proceedings of the 2020 Conference on Empirical Methods in
  Natural Language Processing: System Demonstrations}, pages 119--126.

\bibitem[{Morris et~al.(2020{\natexlab{b}})Morris, Lifland, Lanchantin, Ji, and
  Qi}]{Morris2020ReevaluatingAE}
John~X. Morris, Eli Lifland, Jack Lanchantin, Yangfeng Ji, and Yanjun Qi.
  2020{\natexlab{b}}.
\newblock Reevaluating adversarial examples in natural language.
\newblock In \emph{ArXiv}, volume abs/2004.14174.

\bibitem[{Mrk{\v{s}}i{\'c} et~al.(2016)Mrk{\v{s}}i{\'c}, S{\'e}aghdha, Thomson,
  Ga{\v{s}}i{\'c}, Rojas-Barahona, Su, Vandyke, Wen, and
  Young}]{mrkvsic2016counter}
Nikola Mrk{\v{s}}i{\'c}, Diarmuid~O S{\'e}aghdha, Blaise Thomson, Milica
  Ga{\v{s}}i{\'c}, Lina Rojas-Barahona, Pei-Hao Su, David Vandyke, Tsung-Hsien
  Wen, and Steve Young. 2016.
\newblock Counter-fitting word vectors to linguistic constraints.
\newblock \emph{arXiv preprint arXiv:1603.00892}.

\bibitem[{Nie et~al.(2022)Nie, Guo, Huang, Xiao, Vahdat, and
  Anandkumar}]{DBLP:conf/icml/NieGHXVA22}
Weili Nie, Brandon Guo, Yujia Huang, Chaowei Xiao, Arash Vahdat, and Animashree
  Anandkumar. 2022.
\newblock \href {https://proceedings.mlr.press/v162/nie22a.html} {Diffusion
  models for adversarial purification}.
\newblock In \emph{International Conference on Machine Learning, {ICML} 2022,
  17-23 July 2022, Baltimore, Maryland, {USA}}, volume 162 of \emph{Proceedings
  of Machine Learning Research}, pages 16805--16827. {PMLR}.

\bibitem[{Ren et~al.(2019)Ren, Deng, He, and Che}]{ren2019generating}
Shuhuai Ren, Yihe Deng, Kun He, and Wanxiang Che. 2019.
\newblock Generating natural language adversarial examples through probability
  weighted word saliency.
\newblock In \emph{Proceedings of the 57th Annual Meeting of the Association
  for Computational Linguistics}, pages 1085--1097.

\bibitem[{Samangouei et~al.(2018)Samangouei, Kabkab, and
  Chellappa}]{Samangouei18defense}
Pouya Samangouei, Maya Kabkab, and Rama Chellappa. 2018.
\newblock \href {http://arxiv.org/abs/1805.06605} {Defense-gan: Protecting
  classifiers against adversarial attacks using generative models}.
\newblock \emph{CoRR}, abs/1805.06605.

\bibitem[{Shi et~al.(2021)Shi, Holtz, and Mishne}]{DBLP:conf/iclr/ShiHM21}
Changhao Shi, Chester Holtz, and Gal Mishne. 2021.
\newblock \href {https://openreview.net/forum?id=\_i3ASPp12WS} {Online
  adversarial purification based on self-supervised learning}.
\newblock In \emph{9th International Conference on Learning Representations,
  {ICLR} 2021, Virtual Event, Austria, May 3-7, 2021}. OpenReview.net.

\bibitem[{Shi et~al.(2019)Shi, Huang, Yao, and Xu}]{shi2020robust}
Zhouxing Shi, Minlie Huang, Ting Yao, and Jingfang Xu. 2019.
\newblock \href {http://arxiv.org/abs/1909.02560} {Robustness to modification
  with shared words in paraphrase identification}.
\newblock \emph{CoRR}, abs/1909.02560.

\bibitem[{Si et~al.(2020)Si, Zhang, Qi, Liu, Wang, Liu, and Sun}]{si2020better}
Chenglei Si, Zhengyan Zhang, Fanchao Qi, Zhiyuan Liu, Yasheng Wang, Qun Liu,
  and Maosong Sun. 2020.
\newblock Better robustness by more coverage: Adversarial training with mixup
  augmentation for robust fine-tuning.
\newblock \emph{arXiv preprint arXiv:2012.15699}.

\bibitem[{Song et~al.(2021)Song, Sohl{-}Dickstein, Kingma, Kumar, Ermon, and
  Poole}]{DBLP:conf/iclr/0011SKKEP21}
Yang Song, Jascha Sohl{-}Dickstein, Diederik~P. Kingma, Abhishek Kumar, Stefano
  Ermon, and Ben Poole. 2021.
\newblock \href {https://openreview.net/forum?id=PxTIG12RRHS} {Score-based
  generative modeling through stochastic differential equations}.
\newblock In \emph{9th International Conference on Learning Representations,
  {ICLR} 2021, Virtual Event, Austria, May 3-7, 2021}. OpenReview.net.

\bibitem[{Srinivasan et~al.(2021)Srinivasan, Rohrer, Marb{\'{a}}n,
  M{\"{u}}ller, Samek, and Nakajima}]{DBLP:journals/nn/SrinivasanRMMSN21}
Vignesh Srinivasan, Csaba Rohrer, Arturo Marb{\'{a}}n, Klaus{-}Robert
  M{\"{u}}ller, Wojciech Samek, and Shinichi Nakajima. 2021.
\newblock \href {https://doi.org/10.1016/j.neunet.2020.12.024} {Robustifying
  models against adversarial attacks by langevin dynamics}.
\newblock \emph{Neural Networks}, 137:1--17.

\bibitem[{Wolf et~al.(2020)Wolf, Debut, Sanh, Chaumond, Delangue, Moi, Cistac,
  Rault, Louf, Funtowicz, Davison, Shleifer, von Platen, Ma, Jernite, Plu, Xu,
  Scao, Gugger, Drame, Lhoest, and Rush}]{wolf-etal-2020-transformers}
Thomas Wolf, Lysandre Debut, Victor Sanh, Julien Chaumond, Clement Delangue,
  Anthony Moi, Pierric Cistac, Tim Rault, Rémi Louf, Morgan Funtowicz, Joe
  Davison, Sam Shleifer, Patrick von Platen, Clara Ma, Yacine Jernite, Julien
  Plu, Canwen Xu, Teven~Le Scao, Sylvain Gugger, Mariama Drame, Quentin Lhoest,
  and Alexander~M. Rush. 2020.
\newblock \href {https://www.aclweb.org/anthology/2020.emnlp-demos.6}
  {Transformers: State-of-the-art natural language processing}.
\newblock In \emph{Proceedings of the 2020 Conference on Empirical Methods in
  Natural Language Processing: System Demonstrations}, pages 38--45, Online.
  Association for Computational Linguistics.

\bibitem[{Yoo et~al.(2020)Yoo, Morris, Lifland, and Qi}]{Yoo2020SearchingFA}
Jin~Yong Yoo, John~X. Morris, Eli Lifland, and Yanjun Qi. 2020.
\newblock Searching for a search method: Benchmarking search algorithms for
  generating nlp adversarial examples.
\newblock \emph{ArXiv}, abs/2009.06368.

\bibitem[{Yoon et~al.(2021)Yoon, Hwang, and Lee}]{DBLP:conf/icml/YoonHL21}
Jongmin Yoon, Sung~Ju Hwang, and Juho Lee. 2021.
\newblock \href {http://proceedings.mlr.press/v139/yoon21a.html} {Adversarial
  purification with score-based generative models}.
\newblock In \emph{Proceedings of the 38th International Conference on Machine
  Learning, {ICML} 2021, 18-24 July 2021, Virtual Event}, volume 139 of
  \emph{Proceedings of Machine Learning Research}, pages 12062--12072. {PMLR}.

\bibitem[{Zeng et~al.(2021)Zeng, Zheng, Xu, Li, Yuan, and
  Huang}]{zeng2021certified}
Jiehang Zeng, Xiaoqing Zheng, Jianhan Xu, Linyang Li, Liping Yuan, and Xuanjing
  Huang. 2021.
\newblock Certified robustness to text adversarial attacks by randomized
  [mask].
\newblock \emph{arXiv preprint arXiv:2105.03743}.

\bibitem[{Zhang et~al.(2015)Zhang, Zhao, and LeCun}]{zhang2015character}
Xiang Zhang, Junbo Zhao, and Yann LeCun. 2015.
\newblock Character-level convolutional networks for text classification.
\newblock In \emph{Advances in neural information processing systems}, pages
  649--657.

\bibitem[{Zhou et~al.(2020)Zhou, Zheng, Hsieh, Chang, and
  Huang}]{zhou2020defense}
Yi~Zhou, Xiaoqing Zheng, Cho-Jui Hsieh, Kai-wei Chang, and Xuanjing Huang.
  2020.
\newblock Defense against adversarial attacks in nlp via dirichlet neighborhood
  ensemble.
\newblock \emph{arXiv preprint arXiv:2006.11627}.

\bibitem[{Zhu et~al.(2019)Zhu, Cheng, Gan, Sun, Goldstein, and
  Liu}]{zhu2019freelb}
Chen Zhu, Yu~Cheng, Zhe Gan, Siqi Sun, Thomas Goldstein, and Jingjing Liu.
  2019.
\newblock Freelb: Enhanced adversarial training for language understanding.
\newblock \emph{arXiv preprint arXiv:1909.11764}.

\end{thebibliography}
\bibliographystyle{acl_natbib}

\clearpage

\appendix

\section*{Appendix}

\textbf{Recovery Number Analysis}

One key problem is that how many recoveries we should use in the recovering process, as finding a proper $T$ is also important in the image-domain purification process.
We use two attack methods with $K=12$ to test how the accuracy varies when using different recovery number $N$.

\begin{figure}[h]
\centering
\subfigure[Recovery Number Influence]{
\begin{minipage}[t]{0.95\linewidth}
\includegraphics[width=1\linewidth]{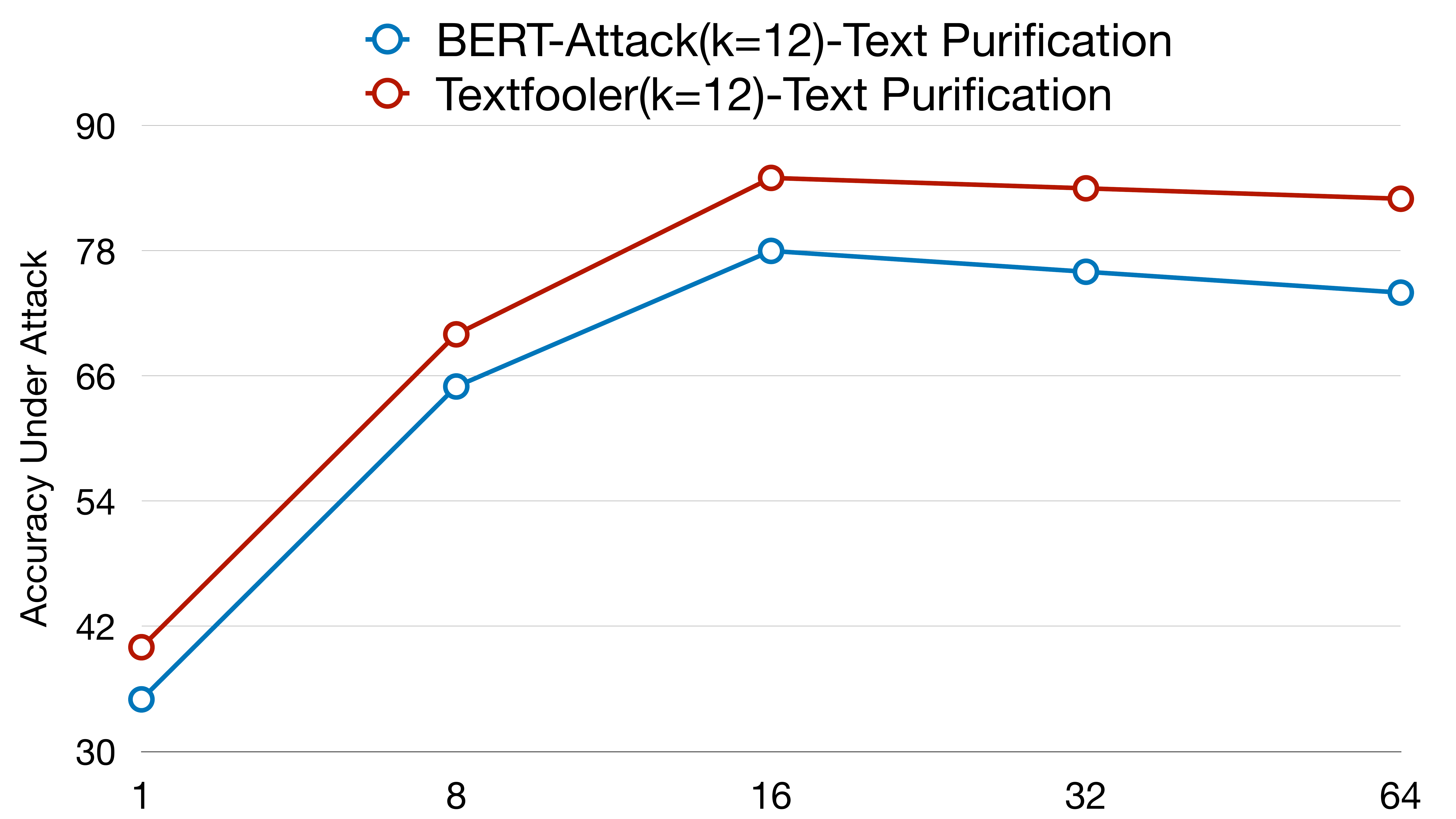}
\end{minipage}%
}%

\subfigure[Candidate-Size Influence]{
\begin{minipage}[t]{0.95\linewidth}
\includegraphics[width=1\linewidth]{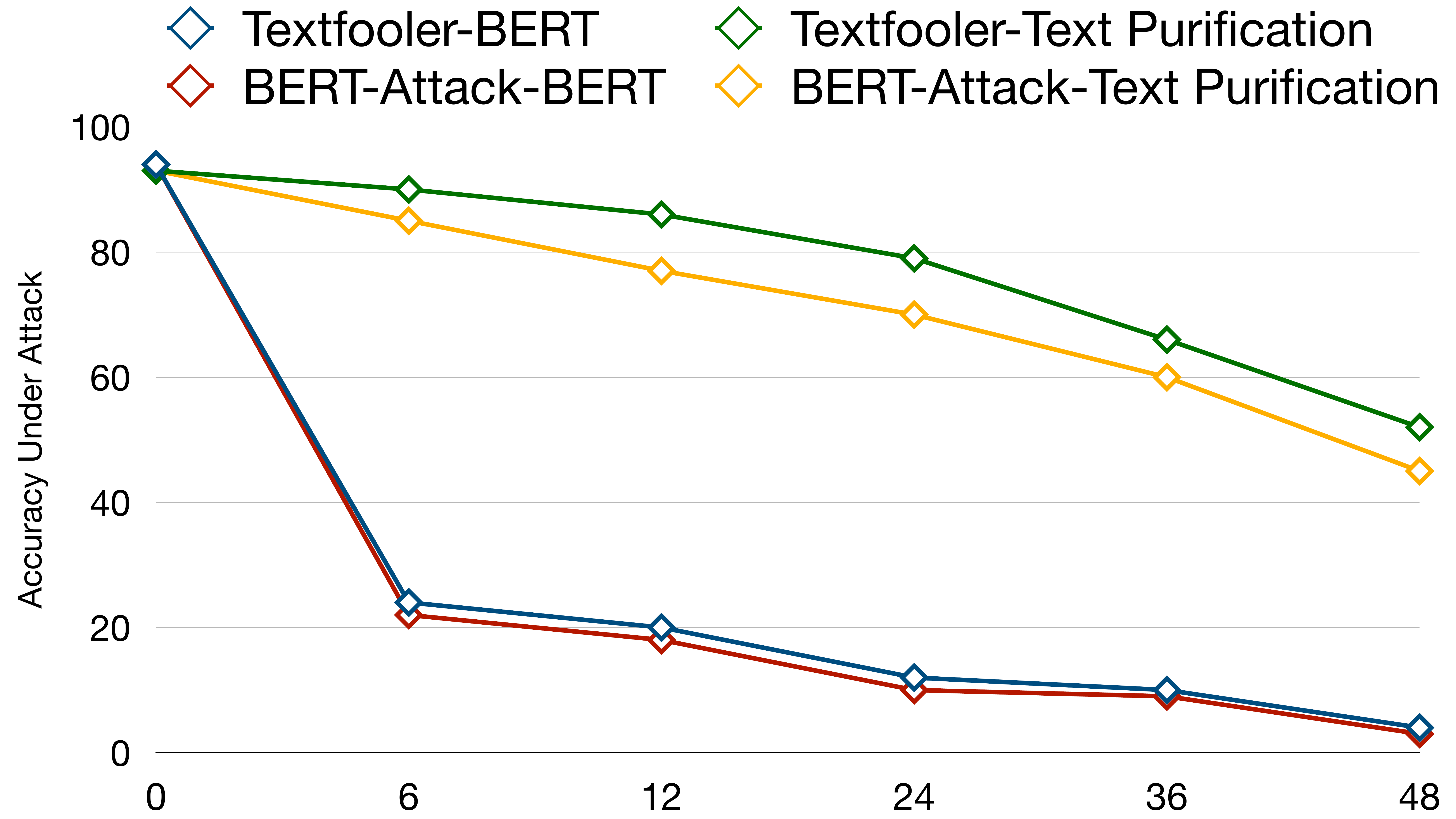}
\end{minipage}%
}%

\centering
\caption{Hyper-Parameter Selection Analysis}
\label{fig:hyper}
\end{figure}

As seen in Fig. \ref{fig:hyper} (a), the ensemble size is actually not a key factor. Larger ensemble size would not result in further improvements. 
We assume that larger ensemble size will \textit{smooth} the output score which will benefit the attack algorithm.
That is, the tiny difference between substitutes can be detected by the attack algorithm since the confidence score is given to the attack algorithms.
Still, we can conclude that a multiple recovery process is effective in the purification process and quite simple to implement.

\textbf{Candidate Size Analysis}

The attack algorithms such as BERT-Attack and Textfooler use a wide range of substitution set (e.g. K=50 in Textfooler means for each token to replace, the algorithm will find the best replacement in 50 candidates), which seriously harms the quality of the input texts.

As seen in Fig. \ref{fig:hyper} (b), when the candidate is 0, the accuracy is high on the clean samples.
When the candidate is 6, the normal fine-tuned BERT model cannot correctly predict the generated adversarial examples.
This indicates that normal fine-tuned BERT is not robust even when the candidate size is small.
After purification, the model can tolerate these limited candidate size attacks. 
When the candidate size grows, the performance of our defense framework drops by a relatively large margin.
We assume that large candidate size would seriously harm the semantics which is also explored in \citet{Morris2020ReevaluatingAE}, while these adversaries cannot be well evaluated even using human-evvaluations since the change rate is still low.

\end{document}